\documentclass[10pt,twocolumn,letterpaper]{article}

\usepackage{iccv}              
\usepackage{multirow}
\usepackage{amsmath,amsfonts}
\usepackage{booktabs,multirow,rotating,array}
\usepackage{makecell}
\usepackage{xcolor}
\usepackage{bbding}
\usepackage{utfsym}
\usepackage{natbib,tabularx}
\usepackage{times}  
\usepackage{helvet} 
\usepackage{courier}  
\usepackage[hyphens]{url} 
\usepackage{graphicx}
\usepackage{colortbl}
\usepackage[ruled,vlined]{algorithm2e}

\definecolor{iccvblue}{rgb}{0.21,0.49,0.74}
\usepackage[pagebackref,breaklinks,colorlinks,allcolors=iccvblue]{hyperref}

\def\paperID{6964} 
\def\confName{ICCV}
\def\confYear{2025}

\title{MGSR: 2D/3D Mutual-boosted Gaussian Splatting for High-fidelity Surface Reconstruction under Various Light Conditions}

\author{
    Qingyuan Zhou\textsuperscript{\rm 1}, Yuehu Gong\textsuperscript{\rm 1}, Weidong Yang\textsuperscript{\rm 1}\thanks{Corresponding authors} , Jiaze Li\textsuperscript{\rm 2}, Yeqi Luo\textsuperscript{\rm 1}, Baixin Xu\textsuperscript{\rm 2}, \\Shuhao Li\textsuperscript{\rm 1}, Ben Fei\textsuperscript{\rm 3}$^*$, Ying He\textsuperscript{\rm 2}\\
    \textsuperscript{\rm 1}Fudan University, 
    \textsuperscript{\rm 2}Nanyang Technological University, \textsuperscript{\rm 3}The Chinese University of Hong Kong\\
    \tt\small zhouqy23@m.fudan.edu.cn, wdyang@fudan.edu.cn, benfei@cuhk.edu.hk, yhe@ntu.edu.sg
}

\begin{document}
\maketitle
\begin{abstract}
Novel view synthesis (NVS) and surface reconstruction (SR) are essential tasks in 3D Gaussian Splatting (3DGS). 
Despite recent progress, these tasks are often addressed independently, with GS-based rendering methods struggling under diverse light conditions and failing to produce accurate surfaces, while GS-based reconstruction methods frequently compromise rendering quality. 
This raises a central question: must rendering and reconstruction always involve a trade-off? 
To address this, we propose \textbf{MGSR}, a 2D/3D \textbf{M}utual-boosted \textbf{G}aussian Splatting for \textbf{S}urface \textbf{R}econstruction that enhances both rendering quality and 3D reconstruction accuracy. 
MGSR introduces two branches—one based on 2DGS and the other on 3DGS. 
The 2DGS branch excels in surface reconstruction, providing precise geometry information to the 3DGS branch. 
Leveraging this geometry, the 3DGS branch employs a geometry-guided illumination decomposition module that captures reflected and transmitted components, enabling realistic rendering under varied light conditions. 
Using the transmitted component as supervision, the 2DGS branch also achieves high-fidelity surface reconstruction. Throughout the optimization process, the 2DGS and 3DGS branches undergo alternating optimization, providing mutual supervision. Prior to this, each branch completes an independent warm-up phase, with an early stopping strategy implemented to reduce computational costs. 
We evaluate MGSR on a diverse set of synthetic and real-world datasets, at both object and scene levels, demonstrating strong performance in rendering and surface reconstruction. 
Code is available at \url{https://github.com/TsingyuanChou/MGSR}.

\end{abstract}

\section{Introduction}
\label{sec:intro}

3D Gaussian Splatting (3DGS)~\cite{kerbl20233d} has recently gained great attention in computer graphics and 3D vision~\cite{fei20243d}.
By representing scenes as collections of 3D Gaussian primitives, 3DGS offers a more flexible and adaptive representation, enabling accurate and efficient rendering and visualization without relying on neural networks.
3DGS effectively resolves the issues of low training efficiency and insufficient geometric accuracy in previous NeRF methods.

Due to its impressive rendering efficiency and accuracy, 3DGS has been widely applied in fields such as SLAM~\cite{matsuki2024gaussian}, dynamic real-world or large-scale scene modeling~\cite{mihajlovic2024splatfields,sabour2024spotlesssplats}, reconstruction~\cite{qian20243dgs,li2024gs},  navigation~\cite{zhou2024drivinggaussian}, manipulation~\cite{qin2024langsplat,lyu2024gaga,liu2024stylegaussian}, 3D generation~\cite{chen2024text,zhou2024gala3d}, and 3D human simulation~\cite{pang2024ash}, etc.

\begin{figure*}[t]
    \centering
    \includegraphics[width=\linewidth]{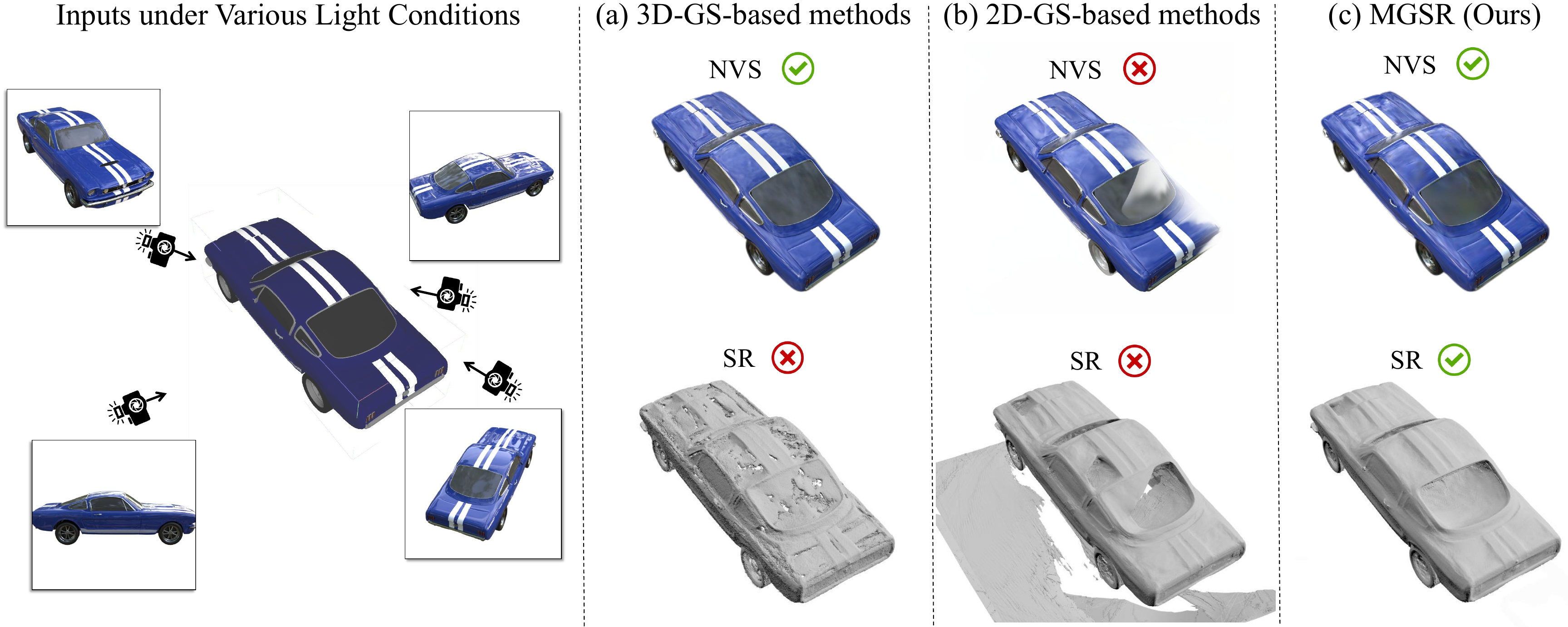}
    \vspace{-0.7cm}
    \caption{MGSR achieves strong NVS and SR results compared with methods based on 2DGS~\cite{huang20242d} and 3DGS~\cite{yu2024gaussian}. 
    The input consists of multi-view images captured from various camera positions and angles, under significantly varying light conditions. In some views, the object appears well-lit and clearly visible, while in others, it is poorly lit, resulting in shadows, reflections, and inconsistencies across the images.}
    \label{fig:teaser}
\vspace{-0.3cm}
\end{figure*}

Nevertheless, achieving high-fidelity surface reconstruction (SR) and improving the realism of novel view synthesis (NVS) under various light conditions are two main challenges in 3DGS.

Specifically, considering that GS-based rendering methods are affected by different light conditions, previous methods~\cite{gao2023relightable,jiang2024gaussianshader} proposed to decompose light to enhance the neural rendering in scenes containing reflective surfaces with the introduction of BRDF (\Cref{fig:teaser}a).
However, despite the effectiveness of illumination decomposition in rendering, these methods are time-consuming and still struggle to achieve meaningful mesh extraction due to inherent limitations in 3DGS.
For instance, the centers of Gaussian primitives do not align well with the surface.
On the other hand, GS-based SR methods~\cite{guedon2024sugar,huang20242d,yu2024gaussian} endeavor to constrain the Gaussian primitives close to the surface, inevitably sacrifice rendering quality and are sensitive to light conditions, which may cause surface artifacts on surfaces (\Cref{fig:teaser}b).
For example, 2DGS~\cite{huang20242d} utilize 2D Gaussian primitives for 3D scene representation, facilitating accurate and view-consistent geometry modeling. However, 2DGS still cannot effectively model surfaces when ambient lighting changes.

To solve these contradictions, we propose \textbf{MGSR}, a 2D/3D \textbf{M}utual-boosted \textbf{G}aussian splatting for \textbf{S}urface \textbf{R}econstruction that enhances both rendering quality and 3D reconstruction accuracy (\Cref{fig:teaser}c). 
Specifically, two branches are devised based on 2DGS and 3DGS, which are optimized synchronously and mutually enhance each other: the 2DGS branch provides accurate geometric information, with its limited rendering capabilities improved by 3DGS branch, while the 3DGS branch focuses on rendering, with the cost of geometric accuracy compensated by the 2DGS branch.
With geometry information, a geometry-guided illumination decomposition module is devised in the 3DGS branch to obtain the reflected and transmitted components, achieving realistic rendering under various light conditions.
To achieve this, an additional Spherical Harmonic (SH) is introduced to model reflected colors, along with two other parameters: reflected opacity and reflected confidence, to represent the reflected component accurately.
Using the transmitted component as supervision, the 2DGS branch can achieve high-fidelity SR while avoiding the impact of illumination factors on surface estimation.
Throughout the mutual-boosted supervision stage, the 2DGS and 3DGS branches engage in alternating optimization for mutual supervision.
Prior to alternating optimization, the two modules undergo an independent warm-up stage, and an auto-stop strategy is introduced to reduce unnecessary computational burdens.
To the best of our knowledge, MGSR is the first GS-based approach that investigates the simultaneous enhancement of rendering and reconstruction, as well as the first mutual-boosted work on GS involving both 2DGS and 3DGS.
MGSR is thoroughly evaluated across a wide range of synthetic and real-world datasets, as well as object- and scene-level datasets, showcasing its superior performance in both rendering and mesh extraction.
In summary, our main contributions are as follows:
\begin{itemize}
\item  The first to explore the feasibility of the joint promotion between rendering and reconstruction, introducing a GS-based 2D/3D mutual-boosted approach that ensures rendering quality while achieving high-fidelity SR under various light conditions.
\item The 2DGS branch aims to provide geometry information to enhance the illumination decomposition of the 3DGS. The decomposed transmitted color will be in turn utilized to supervise the 2DGS branch for improved surface reconstruction, independent of varying light conditions.
\item Leveraging geometric information from the 2DGS branch, we have developed a geometry-guided illumination decomposition module to enhance the accuracy of decomposing reflected and transmitted components, achieving more realistic rendering outcomes.
\item To address the varying convergence speeds in the two branches during the warm-up stage, an auto-stop strategy has been devised. This strategy involves initiating alternating optimization once one branch has completed its warm-up stage.

\end{itemize}
\vspace{-0.3cm}

\section{Related work}
\label{sec:related}

\textbf{Lighting estimation and decomposition.}
Estimating and decomposing light conditions in 3D scenes is a challenging task, further complicated by factors such as reflections, refractions, overexposure, and diverse material properties, resulting in problems related to multi-view inconsistency.
NeRFactor~\cite{zhang2021nerfactor} addresses spatially varying reflectance and environmental lighting using a re-rendering loss, smoothness priors, and data-driven BRDF prior learned from real-world measurements.
NEILF~\cite{yao2022neilf} represents scene lighting as Neural Incident Light and models material properties as surface BRDF. 
NEILF++~\cite{zhang2023neilf++} integrates incident and outgoing light fields through physically-based rendering and surface inter-reflections.
Ref-NeRF~\cite{verbin2022ref} replaces view-dependent emitted radiance parameterization of NeRF with a representation of reflected radiance. 
However, NeRF-based methods are primarily constrained by their significant computational burdens and relatively slow rendering speed.

Recently, several attempts have also been made at lighting estimation and decomposition on 3DGS.
R3DG~\cite{gao2023relightable} leverages 3DGS and NEILF~\cite{yao2022neilf} to create a ray-tracing and relighting-capable 3DGS representation. 
GShader~\cite{jiang2024gaussianshader} presents a simplified shading function for reflective surfaces in 3DGS, 
where a novel normal estimation is introduced that uses the shortest axis direction of the 3D Gaussian as an approximate normal, eliminating reliance on depth map priors and avoiding the flattening or 2D projection of 3D Gaussians.
GS-IR~\cite{liang2024gs} utilizes depth-derivation-based regularization for normal estimation and a baking-based method for modeling indirect lighting, enabling a precise decomposition of material attributes and illumination, thereby significantly improving the photorealism of the rendered outcomes.
The prevailing focus in current efforts on lighting estimation and decomposition is predominantly on achieving photorealistic rendering, which consequently hinders the ability to carry out surface mesh extraction effectively.

\textbf{3DGS driven surface reconstruction.}
\label{subsec:re3}
The misalignment of the centroids (means) of 3D Gaussians with the actual surfaces presents a challenge for accurately reconstructing surfaces.
A common strategy is to flatten the Gaussian spheres~\cite{guedon2024sugar} or utilize 2D Gaussian disks~\cite{huang20242d}, effectively pulling the Gaussian centers closer to the surfaces of the object.
SuGaR~\cite{guedon2024sugar} incorporates a regularization term that enhances the alignment between Gaussians and surfaces within the scene to improve the accuracy of normal estimation. 
2DGS~\cite{huang20242d} adopts flattened 2D Gaussians to represent 3D scenes and defines the normal as the direction of the steepest change in density of the 2D Gaussian distribution. 
GOF~\cite{yu2024gaussian} utilizes ray-tracing-based volume rendering and establishes a level set by opacities of 3D Gaussians.
Surface normals of the Gaussians are approximated using the intersection plane between the ray and the Gaussian, enabling regularization that significantly improves the geometry.
These methods solely focus on scenes with consistent light conditions, which often results in the reconstructed mesh exhibiting significant holes and surface inaccuracies.
Unlike previous methods, PGSR~\cite{chen2024pgsr} takes the light conditions into account and incorporates exposure compensation to enhance the accuracy. 
PGSR flattens the Gaussian into a planar shape and introduces unbiased depth estimation for extracting geometric parameters for surface reconstruction.
However, it may result in over-smoothness in the highlight area, and be challenging in reconstructing reflective or mirror surfaces.
Therefore, to reconstruct meshes under varying light conditions, we propose a method capable of accurate textured mesh extraction in diverse light scenarios.

\section{Method}
\subsection{Overview}
\begin{figure*}[t]
    \centering
    \includegraphics[width=\linewidth]{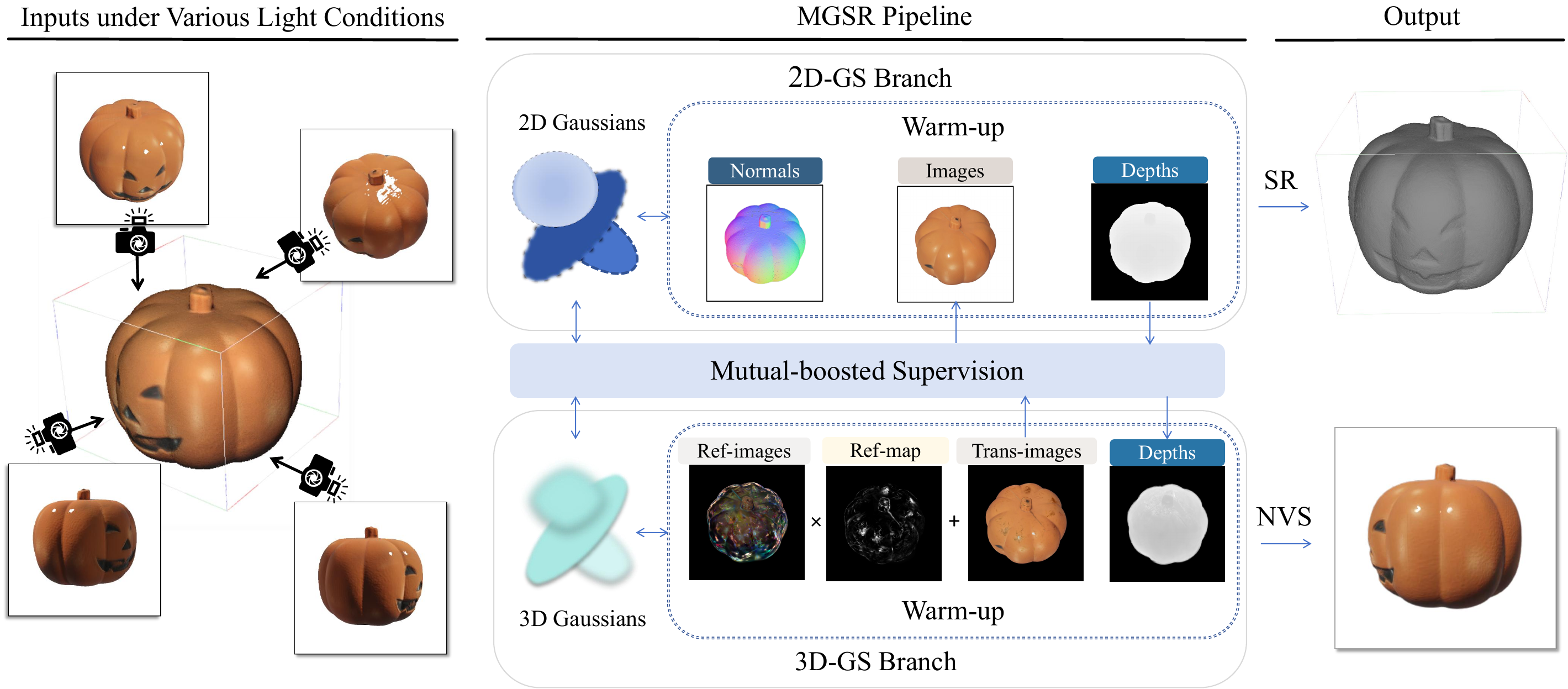}
    \vspace{-0.6cm}
    \caption{
    MGSR is a 2D/3D mutual-boosted framework with two branches: 2DGS branch (upper) for SR and 3DGS branch (bottom) for NVS. Each branch is enhanced by our specific designs. 
    Upon receiving inputs from various light conditions, the two branches initially undergo a warm-up stage of initialization for mutual-boosted optimization. 
    In the alternating optimization process under mutual-boosted supervision, the 3DGS branch is guided by depth maps generated by the 2DGS branch, while simultaneously, rendered transmitted images without reflections from the 3DGS branch provide reflection-removed supervision to the 2DGS branch.}
    \label{fig:pipeline}
\vspace{-0.3cm}
\end{figure*}

MGSR is a 2D/3D mutual-boosted framework that consists of two branches: improved 3DGS branch (\Cref{sec:3dmodule}) and 2DGS branch (\Cref{sec:2dmodule}).
Initially, each branch undergoes an independent warm-up phase, after which they engage in alternating optimization through mutual-boosted supervision (\Cref{sec:2d-3dmodule}).

The 3DGS branch is designed to perform illumination decomposition by introducing reflection-related parameters to disentangle reflections and handle overexposure.
However, accurate illumination decomposition relies on reliable geometry, which the 3DGS branch alone cannot provide. 
To address this limitation, we introduce a geometry-guided illumination decomposition module, which leverages depth information from the 2DGS branch to enhance rendering performance under diverse light conditions. 

Specifically, the 2DGS branch aligns 2D Gaussian disks with the surface, capturing initial and relatively accurate geometry information even under significantly varying light conditions. 
In the subsequent alternating optimization phase, the 3DGS branch is supervised by depth maps generated by the 2DGS branch, while reflection-free images rendered by the 3DGS branch concurrently supervise the 2DGS branch, facilitating a mutual-boosted alternating optimization. 
Finally, following to~\cite{huang20242d}, we employ Truncated Signed Distance Fusion (TSDF)~\cite{zhou2018open3d} to extract the reconstructed textured meshes.

\subsection{Illumination decomposition with 3DGS}
\label{sec:3dmodule}
 
3DGS is constrained in modeling transparent or translucent materials, such as glass.
To enhance the comprehensive modeling of scenes under various light conditions, the entire scene is modeled as composed of transmitted and reflected components. 
Specifically, we retain the original 3DGS as the transmitted component, while introducing three reflection-related parameters to represent the reflected component:
reflected opacity $\theta_{\text{ref}}\in \mathbb{R}$, reflected confidence $\beta \in [0,1]$, and reflected SH $\mathcal{C}_{\text{ref}} \in \mathbb{R}^k$, where $k$ refers to the degrees of freedom.

The reflected confidence $\beta$ represents the probability that an individual Gaussian primitive captures the reflected component. 
When splatting 3D Gaussians to 2D images, $\beta$ is accumulated as described in \Cref{eq:m1} to obtain the pixel-wise reflected confidence $W$, and the transmitted color $C_{\text{tran}}$ and reflected color $C_{\text{ref}}$ are calculated with $\alpha$-blending from front to back, as depicted in:
\begin{equation}\small
    W = \sum_{i=1}^K \beta^i {\alpha_{\text{ref}}^i} \prod_{j=1}^{k-1}\left(1-\beta^j\right),
    \label{eq:m1}
\end{equation}
\begin{equation}\small
    C_{\text{tran}}=\sum_{i=1}^K c_{\text{tran}}^i \alpha_{\text{tran}}^i \prod_{j=1}^{i-1}\left(1-\alpha_{\text{tran}}^j\right),
    \label{eq:m2}
\end{equation}
\begin{equation}\small
    C_{\text{ref}}=\sum_{i=1}^K c_{\text{ref}}^i \alpha_{\text{ref}}^i \prod_{j=1}^{i-1}\left(1-\alpha_{\text{ref}}^j\right),
    \label{eq:m3}
\end{equation}
where $i$ denotes the index of the Gaussian sphere, $\alpha_{\text{tran}}$ and $\alpha_{\text{ref}}$ denote to the transmitted opacity $\theta_{\text{tran}}$ and reflected opacity $\theta_{\text{ref}}$  multiplied by the density of the splatted Gaussian at the pixel location, $c_{\text{tran}}^i$ and $c_{\text{ref}}^i$ represent the view-dependent transmitted and reflected color calculated from transmitted SH $\mathcal{C}_{\text{tran}}$ and reflected SH $\mathcal{C}_{\text{ref}}$. 
The rendered color $C$ is calculated according to \Cref{eq:m4}, combining the transmitted and reflected components, weighted by the reflected confidence $W$:
\begin{equation}\small
C = C_{\text{tran}} + W \times C_{\text{ref}}.
\label{eq:m4}
\end{equation}
A total variation (TV) loss $\mathcal{L}_{\text{trans-TV}}$ is utilized to smooth in local regions of the transmitted components.
Subsequently, the rendering loss is applied to encourage rendered color $C$ to be similar to the GT color $C_{\text{GT}}$.
\begin{equation}\small
    \mathcal{L}_{\text{render}} = \lambda_1 \mathcal{L}_1(C, C_{\text{GT}}) + (1 - \lambda_1) \mathcal{L}_{\text{D-SSIM}}(C, C_{\text{GT}}),
\label{eq:m5}
\end{equation}
where $\lambda_1$ represents to the balance coefficient, $\mathcal{L}_1$ computes the absolute error, while $\mathcal{L}_{\text{D-SSIM}}$ refers to the differentiable structural similarity index measure (SSIM).
The total loss is a weighted sum of the rendering loss and the TV loss of transmitted components, 
\begin{equation}\small
\mathcal{L}_{\text{3D}} = \mathcal{L}_{\text{render}} + \lambda_2 \mathcal{L}_{\text{trans-TV}},
\label{eq:m6}
\end{equation}
where $\lambda_2$ denotes to the weight.

However, due to the inherent limitation of 3DGS, specifically the inaccuracy of the depth map, although this 3D branch successfully performs illumination decomposition, it still cannot extract the surface mesh. 
Therefore, an additional 2DGS branch is introduced to provide reliable geometry supervision.

\subsection{Surface reconstruction with 2DGS}\label{sec:2dmodule}

2DGS represents the scene with flattened 2D Gaussian primitives, aligning the centers of the Gaussian disks with the surface.
The aim of this section is to utilize these flattened Gaussian primitives to obtain initialized depth maps of scenes under various light conditions, which also serve as supervision for the 3DGS branch and are refined iteratively through the mutual-boosted stage (\Cref{sec:2d-3dmodule}).

As an initial estimate, the rendering depth map $Z$ is computed as a weighted sum of the normalized intersected depths $z$, as depicted in:
\begin{equation}\small
Z = \frac{\sum_{i} \omega_i z_i}{\sum_{i} \omega_i + \epsilon},
\label{eq:m8}
\end{equation}
where $\omega_i = T_i \alpha_i$ is the contribution of 2D Gaussian disk to the rendering depth at the pixel location.
$\alpha_i$ is the opacity $\theta$ multiplied by the density of the splatted Gaussian, and the visibility term $T_i$ is calculated:
\begin{equation}\small
T_i = \prod_{j=1}^{i-1}\left(1-\alpha_j\right),
\label{eq:m9}
\end{equation}
where the definition of $\alpha_j$ is same as above.

In the representation based on 2D Gaussian primitives, it is crucial to ensure that all 2D splats are locally aligned with the actual surfaces of the object. 
This alignment is particularly important in the context of volume rendering, where multiple semi-transparent surfels may be encountered along the ray. 
To accurately identify the actual surface, we follow~\cite{huang20242d} and consider the median point of intersection where the accumulated opacity reaches 0.5.
The normals of the 2D splats are encouraged to be aligned with the gradients of the depth maps, as shown in:
\begin{equation}
\mathcal{L}_{\text{n}} = \sum_{i} \omega_i (1 - n_i^\top N),
\label{eq:m10}
\end{equation}
where $i$ indexes the intersected splats along the ray, $n_i$ represents the normal of the splat oriented towards the camera, and $N$ is the normal estimated from the nearby depth point $Z$, and computed by:
\begin{equation}
N(x, y) = \frac{\nabla_x Z \times \nabla_y Z}{\|\nabla_x Z \times \nabla_y Z\|}.
\label{eq:m11}
\end{equation}

Considering that our task is to reconstruct the surface meshes under various light conditions, a weighted normal loss is employed, which encourages the model to primarily learn the normals of the surfaces rather than those of the entire scenes.
The normal loss from the camera's perspective will be determined based on the division of the mask into foreground and background, with a coefficient $\gamma$ applied to weight the contributions.
Two TV losses $\mathcal{L}_{\text{d-TV}}$ and $\mathcal{L}_{\text{n-TV}}$ are introduced as smooth terms on both rendered depths and normals. The overall loss of the 2DGS branch consists of a weighted combination:
\begin{equation}
\mathcal{L}_{\text{2D}} = \mathcal{L}_{\text{render}}+\lambda_3(\gamma \mathcal{L}_{\text{n}}+\lambda_4\mathcal{L}_{\text{n-TV}}) +\lambda_5\mathcal{L}_{\text{d-TV}},
\label{eq:m11_2}
\end{equation}
where $\lambda_3$, $\lambda_4$, and $\lambda_5$ represent the weights, and $\mathcal{L}_{\text{render}}$ is defined in \Cref{eq:m5}.

Yet, the 2DGS branch has only accomplished modeling of illuminated scenes and relatively reliable depth estimation without eliminating the influence of light conditions, which is inconsistent with real unlit scenes.
To address this issue, an alternating optimization approach is devised after the warm-up stage of both the 2DGS and 3DGS branches.

\subsection{Alternating optimization of 2D \& 3D Gaussians }\label{sec:2d-3dmodule}

\begin{figure}[t]
    \centering
    \includegraphics[width=\linewidth]{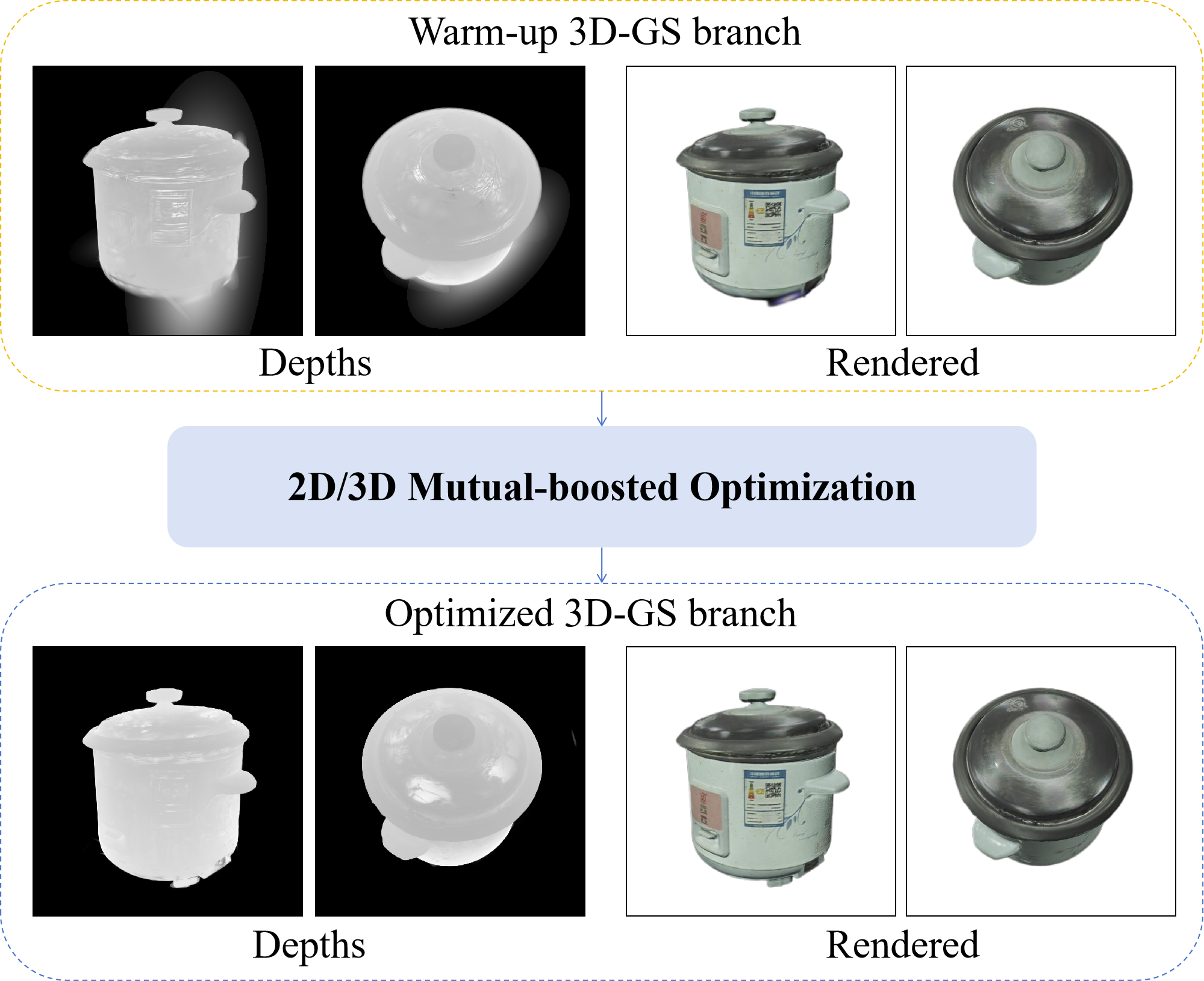}
    \vspace{-0.3cm}
    \caption{Geometry enhancement in 3DGS branch for realistic rendering through our mutual-boosted optimization.}
    \label{fig:branch}
\vspace{-0.4cm}
\end{figure}

Given that the warm-up 3DGS branch has performed a preliminary illumination decomposition in \Cref{sec:3dmodule}, and the warm-up 2DGS branch has provided an initial estimate of the depth maps for the scenes under various light conditions in \Cref{sec:2dmodule}, our alternating optimization approach utilizes the transmitted colors from the 3DGS branch to supervise the 2DGS branch for surface reconstruction, while the depth maps from the 2DGS branch provides geometry information for the 3DGS branch for better illumination decomposition. 

Specifically, since the illumination decomposition in the 3DGS branch is unsupervised, the transmitted components from 3DGS branch might not be completely disentangled from the reflected components after the warm-up stage, a weighted rendering loss is introduced for rendering supervision for the 2DGS branch.
Our goal is for the images rendered by the 2DGS branch to not only approximate the transmitted images but also to retain a slight similarity to the illuminated images to enhance the stability of alternating optimization, as calculated by:
\begin{equation}
 \mathcal{L}_{\text{render-m}} = \lambda_6 \mathcal{L}_{\text{mutual}} +(1-\lambda_6)\mathcal{L}_{\text{2D-render}},
\label{eq:m12}
\end{equation}
where $\mathcal{L}_{\text{render-m}}$ represents the mutual rendering loss in alternating optimization, and $\lambda_6$ is the weight. 
$\mathcal{L}_{\text{mutual}}$ denotes the rendering loss between the 2DGS branch rendered images and the transmitted images from the 3DGS branch, $\mathcal{L}_{\text{2D-render}}$ is the rendering loss between the 2DGS branch rendered images and the GT images, and both of the rendering loss are computed by \Cref{eq:m5}.
Moreover, both TV losses on depths ($\mathcal{L}_{\text{d-TV-m}}$) and normals ($\mathcal{L}_{\text{n-TV-m}}$) in 2DGS branch are retained, but GT images are replaced with the transmitted images.
In the alternating optimization stage, the loss function of 2DGS branch will be promoted to:
\begin{equation}
\mathcal{L}_{\text{2D}} = \mathcal{L}_{\text{render-m}}+\lambda_3(\gamma \mathcal{L}_{\text{n}}+\lambda_4\mathcal{L}_{\text{n-TV-m}}) +\lambda_5\mathcal{L}_{\text{d-TV-m}},
\label{eq:m2-1}
\end{equation}
where $\gamma$ is the coefficient that balances the contribution of the foreground and background. The weights $\lambda_2$, $\lambda_3$, and $\lambda_4$ are the same as the warm-up stage.

\begin{figure*}[h]
    \centering
    \includegraphics[width=\linewidth]{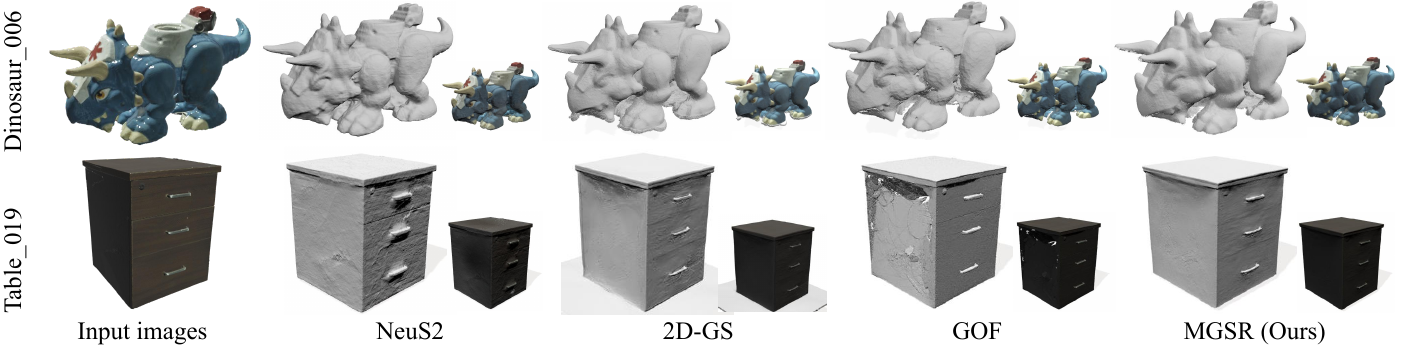}
    \vspace{-0.8cm}
    \caption{Visual comparisons on the OmniObject3D dataset~\cite{wu2023omniobject3d}.}
    \label{fig:exp_oo3d}
\vspace{-0.2cm}
\end{figure*}

\begin{figure*}[h]
    \centering
    \includegraphics[width=\linewidth]{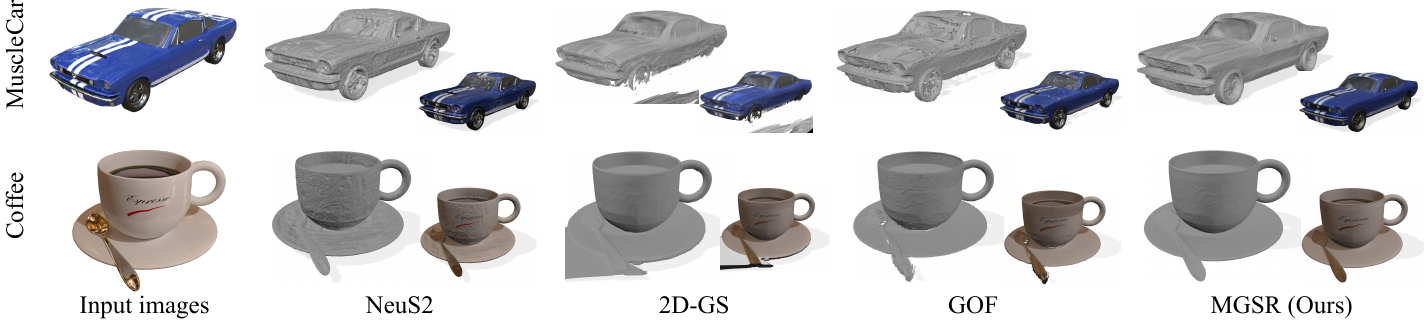}
    \vspace{-0.8cm}
    \caption{Visual comparisons on the Shiny Blender dataset~\cite{verbin2022ref}.}
    \label{fig:exp_Shiny}
\vspace{-0.6cm}
\end{figure*}

Scene geometry is essential for realistic physically-based rendering, and geometry estimation on Gaussian primitives is difficult due to the discrete structures.
Therefore, on the other hand, a depth loss between 2DGS and 3DGS branches is introduced to improve the geometry estimation for the illumination decomposition of 3DGS branch (as shown in \Cref{fig:branch}). 
The calculation of depth maps $Z_{\text{3D}}$ in 3DGS branch is the same as \Cref{eq:m8}. 
To encourage the alternating optimization to focus on the foreground part of the input images, masks are used to avoid the calculation of background by with a weight coefficient $\gamma$. 
The depth loss $\mathcal{L}_{Z}$ is computed by:
\begin{equation}
\mathcal{L}_{Z} = \gamma \mathcal{L}_2(Z_{\text{2D}},Z_{\text{3D}}),
\label{eq:m13}
\end{equation}
where $\mathcal{L}_2$ denotes the L2 loss, and $Z_{\text{2D}}$ is the estimated depth map $Z$ in the 2DGS branch.

The total loss $\mathcal{L}_{\text{total}}$ of the alternating optimization is:
\begin{equation}
\mathcal{L}_{\text{total}} = w_{\text{2D}}\mathcal{L}_{\text{2D}} + w_{\text{3D}}\mathcal{L}_{\text{3D}} + w_{\text{depth-mutual}}\mathcal{L}_{Z},
\label{eq:m14}
\end{equation}
where the losses of the 3D module $\mathcal{L}_{\text{3D}}$ and the 2D module $\mathcal{L}_{\text{2D}}$ are calculated by \Cref{eq:m6} and   \Cref{eq:m2-1}, respectively, and $w_{\text{2D}}$, $w_{\text{3D}}$, and $w_{\text{depth-mutual}}$ are the weights. See \Cref{sec:ablation} for their settings.

\textbf{Auto-stop strategy.} Moreover, due to the varying convergence speeds in each branch, we should not warm up both branches simultaneously at the same iterations.
Therefore, to prevent the two branches from being overly optimized during the warm-up stage and to expedite the overall process, an auto-stop strategy is introduced in the warm-up process. 
Specifically, as one branch reaches convergence, it will initiate our alternating optimization process first. Subsequently, when the other branch also converges, our mutual-boosted stage starts.

\section{Experiments}
\label{sec:exp}

\begin{figure*}[h]
    \centering
    \includegraphics[width=\linewidth]{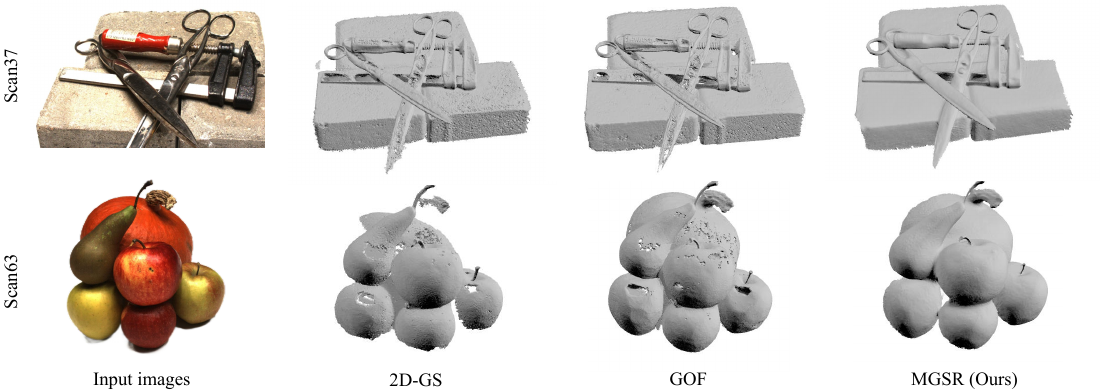}
    \vspace{-0.7cm}
    \caption{Visual comparisons on DTU dataset~\cite{jensen2014large}.}
    \label{fig:exp_dtu}
\vspace{-0.4cm}
\end{figure*}

\begin{figure*}[h]
    \centering
    \includegraphics[width=\linewidth]{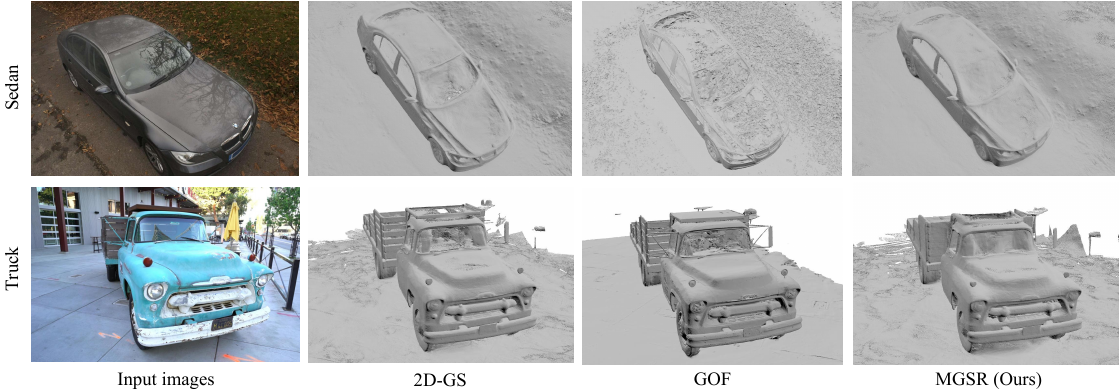}
    \vspace{-0.7cm}
    \caption{Visual comparisons on Ref-NeRF Real Captured Scenes dataset (Sedan)~\cite{verbin2022ref} and TnT dataset (Truck)~\cite{jensen2014large}.}
    \label{fig:exp_refreal}
\vspace{-0.6cm}
\end{figure*}

\subsection{Datasets and evaluation metrics}
\textbf{DTU}~\cite{jensen2014large} is a large MVS dataset, where some scenes feature unfavorable light conditions for surface reconstruction, such as overexposure, underexposure, and metallic reflections. 
Similar to previous baselines~\cite{huang20242d,yu2024gaussian}, we utilized the same 15 scans from the DTU dataset to validate our approach.
\textbf{OmniObject3D}~\cite{wu2023omniobject3d} contains objects with extensive areas of specular highlights on their surfaces.
\textbf{Shiny Blender}~\cite{verbin2022ref} is introduced to assess the capability of surface mesh extraction under strong reflections. 
\textbf{Ref-NeRF Real Captured Scenes}~\cite{verbin2022ref} consists of three in-the-wild scenes with strong reflections. 
\textbf{Tanks and Temples (TnT)}~\cite{knapitsch2017tanks} consists a large number of high-resolution real-world images under varying light conditions.

\begin{table}[t]
\caption{NVS results on Shiny Blender and OmniObject3D. The instance-level metrics are listed in Appendix.}
\vspace{-0.3cm}
\centering
\resizebox{\linewidth}{!}{
\begin{tabular}{c|cc|ccc}
\toprule [1.5pt]
\multirow{2}{*}{Methods}&\multicolumn{2}{c|}{Shiny Blender}&\multicolumn{3}{c}{OmniObject3D}\\
&SSIM$\uparrow$ &PSNR$\uparrow$ &SSIM$\uparrow$ &PSNR$\uparrow$&Time$\downarrow$\\ \toprule [1pt]
3D-GS&0.9630&32.62&\underline{0.9859}&33.64   &\textbf{7min}\\
GS-IR&0.9409&32.09&0.9752&35.16&33min\\
GShader&\textbf{0.9679}&\underline{34.35}&0.9840&34.14&105min\\
R3DG&\underline{0.9646}&34.20&\textbf{0.9874}&\underline{37.37}&21min\\
\toprule [1pt]
MGSR (Ours)&0.9645&\textbf{34.66}&0.9850&\textbf{37.69}  &\underline{13min}\\
\toprule [1.5pt]
\end{tabular}
}
\label{table:new_exp_nvs}
\vspace{-0.5cm}
\end{table}
\begin{table}[t]
\caption{SR results on Shiny Blender and OmniObject3D. NC is multiplied by $10^{2}$, and CD is multiplied by $10^{3}$. Methods marked with {*} fail on certain objects, which are excluded from the average metric values presented in this table, but are detailed in the list of instance-level metrics in the Appendix.}
\vspace{-0.3cm}
\centering
\resizebox{\linewidth}{!}{
\begin{tabular}{c|cccc|ccccc}
\toprule [1.5pt]
\multirow{2}{*}{Methods}&\multicolumn{4}{c|}{Shiny Blender}&\multicolumn{5}{c}{OmniObject3D}\\
&NC$\uparrow$ &CD$\downarrow$ &SSIM$\uparrow$ &PSNR$\uparrow$&NC$\uparrow$ &CD$\downarrow$ &SSIM$\uparrow$ &PSNR$\uparrow$&Time$\downarrow$\\ \toprule [1pt]
NeuS2&62.58&\textbf{0.51}&-&-&88.21&\textbf{0.48}&-&-&\textbf{5min}\\
2D-GS{*}&\underline{65.48}&41.54&0.9597&33.46&86.34&\underline{0.75}&0.9836&34.09   &\underline{12min}\\
GOF{*}&44.54&\underline{2.11}&\textbf{0.9667}&\textbf{34.66}&\underline{90.06}&0.84&\textbf{0.9885}&\textbf{38.18}
&16min \\\toprule [1pt]
MGSR (Ours)&\textbf{66.47}&2.85&\underline{0.9645}&\textbf{34.66}&\textbf{90.60}&0.90&\underline{0.9850}&\underline{37.69}  &13min\\
\toprule [1.5pt]
\end{tabular}
}
\label{table:new_exp_sr}
\vspace{-0.8cm}
\end{table}

\begin{table*}[h]
\caption{Instance-level Normal Consistency results of SR on DTU dataset. Results are multiplied by $10^3$.}
\vspace{-0.3cm}
\centering
\resizebox{\linewidth}{!}{
\begin{tabular}{c|ccccccccccccccc|c}
\toprule [1.5pt]
Methods&24&37&40&55&63&65&69&83&97&105&106&110&114&118&122&Mean\\\toprule [1pt]
2D-GS&50.58&\textbf{49.89}&45.73&46.09&\textbf{68.10}&\textbf{54.62}&54.36&54.72&60.33&33.12&\textbf{49.59}&71.81&60.83&44.95&47.00&52.78\\
GOF&38.13&47.76&45.47&41.17&61.47&49.84&52.35&\textbf{55.00}&55.85&42.75&48.70&73.11&55.65&52.01&\textbf{47.05}&51.09\\
MGSR (Ours)&\textbf{52.46}&48.59&\textbf{47.47}&\textbf{51.66}&55.17&45.34&\textbf{63.27}&40.71&\textbf{61.51}&\textbf{52.40}&42.59&\textbf{74.08}&\textbf{65.04}&\textbf{52.16}&43.81&\textbf{53.08}\\
\toprule [1.5pt]
\end{tabular}
}
\label{tb:exp_dtu}
\vspace{-0.4cm}
\end{table*}
\begin{table*}[h]
\caption{Ablations of loss weights (Models A-F), iterations of mutual-boosted optimization (Models G-J), bidrectional BP and auto-stop warm-up strategy (Models K-L) on OmniObject3D dataset.
NC is multiplied by $10^{2}$, and CD is multiplied by $10^{3}$.
The best-performing model in each ablation study is highlighted.}
\centering
\vspace{-0.3cm}
\resizebox{\linewidth}{!}
{
\begin{tabular}{c|c|c|c|ccc|cccc}
\toprule [1.5pt]
Model&Mutual-boosted Iterations&Bidirectional BP&Auto-stop Warm-up&$w_{\text{2D}}$ &$w_{\text{3D}}$& $w_{\text{depth-mutual}}$&SSIM$\uparrow$&PSNR$\uparrow$&NC$\uparrow$&CD$\downarrow$\\ \toprule [1pt]
A&20k&No&Yes&0&1&0&0.9755&35.10&-&-\\
B&20k&No&Yes&1&0&0&-&-&90.16&0.92\\
C&20k&No&Yes&0.3&0.7&0.01&\textbf{0.9861}&\textbf{37.88}&90.54&0.92\\
D&20k&No&Yes&0.7&0.3&0.01&0.9825&36.99&\textbf{90.62}&0.92\\
\cellcolor{pink}E&\cellcolor{pink}20k&\cellcolor{pink}No&\cellcolor{pink}Yes&\cellcolor{pink}0.5&\cellcolor{pink}0.5&\cellcolor{pink}0.01&\cellcolor{pink}\underline{0.9850}&\cellcolor{pink}\underline{37.69}&\cellcolor{pink}\underline{90.60}&\cellcolor{pink}\textbf{0.90}\\
F&20k&No&Yes&0.5&0.5&0.1&0.9803&36.42&\underline{90.60}&\underline{0.91}\\\toprule [1pt]
G&5k&No&Yes&0.5&0.5&0.01&0.9820&36.76&90.40&\textbf{0.89}\\
H&10k&No&Yes&0.5&0.5&0.01&0.9842&37.37&90.50&0.91\\
\cellcolor{pink}I&\cellcolor{pink}20k&\cellcolor{pink}No&\cellcolor{pink}Yes&\cellcolor{pink}0.5&\cellcolor{pink}0.5&\cellcolor{pink}0.01&\cellcolor{pink}\textbf{0.9850}&\cellcolor{pink}\underline{37.69}&\cellcolor{pink}\textbf{90.60}&\cellcolor{pink}\underline{0.90}\\
J&30k&No&Yes&0.5&0.5&0.01&\textbf{0.9850}&\textbf{37.71}&\textbf{90.60}&0.91\\
\toprule [1pt]
K&20k&Yes&Yes&0.5&0.5&0.01&0.9813&36.64&90.57&0.91\\
L&20k&No&No&0.5&0.5&0.01&0.9831&37.01&90.56&0.92\\
\cellcolor{pink}M&\cellcolor{pink}20k&\cellcolor{pink}No&\cellcolor{pink}Yes&\cellcolor{pink}0.5&\cellcolor{pink}0.5&\cellcolor{pink}0.01&\cellcolor{pink}\textbf{0.9850}&\cellcolor{pink}\textbf{37.69}&\cellcolor{pink}\textbf{90.60}&\cellcolor{pink}\textbf{0.90}\\
\toprule [1.5pt]
\end{tabular}
}
\label{tab:ab}
\vspace{-0.4cm}
\end{table*}
We utilize SSIM and PSNR to evaluate the rendering quality, while reconstruction accuracy is validated by 10K sampled points with Normal Consistency (NC) and Chamfer Distance (CD) measurements.
Due to the limitation of CD, we mainly focus on NC metric, which aligns better with human perception.
Since GT mesh is unavailable for real-world data, only visual comparisons are provided.

\subsection{Results}\label{sec:result}
\textbf{Object-level data.} 
Three objects from the Shiny Blender with reflections and 30 objects from OmniObject3D with highlights are conducted on all methods for comparison.
\Cref{table:new_exp_nvs} and \Cref{table:new_exp_sr} present quantitative results on both NVS and SR.
MGSR surpasses all baselines in terms of the PSNR metric for NVS and the NC metric for SR.
As shown in visualization comparisons of \Cref{fig:exp_oo3d} and \Cref{fig:exp_Shiny}, although NeuS2 exhibits superior CDs, its smoothness on surfaces with reflections is significantly poor.
To address this issue, NC is introduced as an evaluation metric for reconstruction, overcoming the limitations of CDs, which fail to capture surface holes or bumps.
2DGS produces excessive faces on reconstructed meshes on the exterior of objects while failing to accurately reconstruct thin surfaces.
GOF tends to reconstruct rough or incomplete surfaces when handling lit surfaces. 
MGSR visually outperforms all baselines, resulting in the best NC, with smooth surfaces and accurate color modeling. 
\Cref{tb:exp_dtu} and \Cref{fig:exp_dtu} show the quantitative and visual results compared to GS-based SR methods on DTU dataset.
Under varying light conditions, MGSR successfully reconstructs realistic and intact surfaces compared to 2DGS and GOF.

\textbf{Scene-level data.} As shown in \Cref{fig:exp_refreal}, two in-the-wild scenes from Ref-NeRF~\cite{verbin2022ref} and TnT~\cite{knapitsch2017tanks} are used for visual comparisons.
More comparisons are shown in the Appendix.
Previous GS-based methods fail to effectively reconstruct glass or mirror surfaces, resulting in damaged and inaccurate surfaces. 
In contrast, MGSR successfully reconstructs meshes in mirror material, driven by our devised 3DGS branch and mutual-boosted optimization, which perform illumination decomposition and guide the 2DGS branch for enhanced SR.

\textbf{Optimization time.} 
In \Cref{table:new_exp_nvs} and \Cref{table:new_exp_sr}, we additionally present the optimization time for all compared methods. 
By eliminating the dependence on BRDF, MGSR is faster than all illumination decomposition baselines.
Among methods supporting both NVS and SR, 2DGS is the fastest. 
MGSR, supported by an auto-stop warm-up strategy, outperforms GOF and achieves a comparable speed with 2DGS.
Specifically, each of the two branches in MGSR has an average warm-up optimization time of around 3.5-4 minutes.
The mutual-boosted optimization time is approximately 8-9 minutes.

\section{Ablation studies}
\label{sec:ablation}

All ablation studies (\Cref{tab:ab}) are conducted on the OmniObject3D dataset due to its various light conditions, with the experimental setup consistent with \Cref{sec:exp}. 
Additional results are provided in the Appendix.

\textbf{Auto-stop warm-up.} 
Experimental results (Models G-J) show that the auto-stop strategy during warm-up outperforms both MVS and SR, demonstrating the effectiveness of the auto-stop strategy in preventing overfitting during the warm-up stage.
\textbf{Bidirectional back-propagation (BP).} 
In alternating optimization (\Cref{sec:2d-3dmodule}), the outputs from one branch are utilized to supervise the other branch.
During the supervision, the gradient of loss can be back-propagated to one branch (unidirectional BP) or both branches (bidirectional BP).
Experimental results (Models K and M) show that unidirectional BP outperforms bidirectional BP in terms of both NVS and SR results.
It is mainly due to the bidirectional BP influences optimization of the branch which provides supervision.
\textbf{Mutual-boosted iterations.} 
We observed that when the mutual-boosted iterations exceed 20k, there is no significant improvement in the NVS and SR metrics (Models L and M). Therefore, 20k iterations are optimal for MGSR.
\textbf{Loss weights.} 
The cases without mutual-boosted optimization are firstly validated, where only the 3DGS branch is used for NVS or the 2DGS branch for SR (Models A and B). 
The results were inferior compared to the case with mutual-boosted optimization. 
Furthermore, it is found that the branch with a higher weight (Models C and D) during the mutual-boosted optimization stage tends to favor the task it excels at, leading to better performance.
To balance both NVS and SR, the weights of the two branches should be equal.


\vspace{-0.08cm}
\section{Conclusion}
\label{sec:con}

We introduce MGSR, a 2D/3D Mutual-boosted Gaussian splatting for SR that enhances both rendering quality and 3D reconstruction accuracy under various light conditions. 
Moreover, the auto-stop strategy is proposed in the warm-up stage, while geometry-guided illumination decomposition is devised in the mutual-boosted optimization stage.
Extensive experiments across various synthetic and real-world datasets at both object and scene levels have validated the superiority of MGSR.

\textbf{Limitations and future work.} 
Although MGSR achieves reliable reconstruction under reflective conditions, over-smoothness may occur in low-light scenarios, leading to the absence of details. A possible way for addressing this issue is to incorporate exposure compensation for input images, which we will investigate as a future work.

\newpage
\clearpage
{\section*{Acknowledgment}
This project was partially supported by the Ministry of Education, Singapore, under its Academic Research Fund Grant (RT19/22). The computational experiments were conducted using the CFFF platform at Fudan University.
    \small
    \bibliographystyle{ieeenat_fullname}
    \bibliography{main}
}

\end{document}